# Accurate Detection of Inner Ears in Head CTs Using a Deep Volume-to-Volume Regression Network with False Positive Suppression and a Shape-Based Constraint

Dongqing Zhang, Jianing Wang, Jack H. Noble and Benoit M. Dawant

Department of Electrical Engineering and Computer Science, Vanderbilt University, Nashville, TN, 37235, USA
dongqing.zhang@vanderbilt.edu

**Abstract.** Cochlear implants (CIs) are neural prosthetics which are used to treat patients with hearing loss. CIs use an array of electrodes which are surgically inserted into the cochlea to stimulate the auditory nerve endings. After surgery, CIs need to be programmed. Studies have shown that the spatial relationship between the intra-cochlear anatomy and electrodes derived from medical images can guide CI programming and lead to significant improvement in hearing outcomes. However, clinical head CT images are usually obtained from scanners of different brands with different protocols. The field of view thus varies greatly and visual inspection is needed to document their content prior to applying algorithms for electrode localization and intra-cochlear anatomy segmentation. In this work, to determine the presence/absence of inner ears and to accurately localize them in head CTs, we use a volume-to-volume convolutional neural network which can be trained end-to-end to map a raw CT volume to probability maps which indicate inner ear positions. We incorporate a false positive suppression strategy in training and apply a shape-based constraint. We achieve a labeling accuracy of 98.59% and a localization error of 2.45mm. The localization error is significantly smaller than a random forest-based approach that has been proposed recently to perform the same task.

## 1 Introduction

Cochlear implants (CIs) have been one of the most successful neural prosthetics in the past few decades [1]. They are used to treat patients with severe to profound hearing loss. During a cochlear implantation surgery, an array of electrodes is threaded into the cochlea to replace the natural signal transduction mechanism in human hearing. After surgery, the CI needs to be programmed to adjust the implant for the recipient. The programming includes the assignment of a frequency range to each individual contact in the array so it will be activated when the incoming sound includes frequency components in this range. Traditionally, the programming is done by an audiologist who can only rely on the recipients' subjective response to certain stimuli, e.g., whether they can hear a signal or rank pitches, without other clues. Accurately localizing electrodes in CI relative to the intra-cochlear anatomy can provide useful guidance to audiologists to adjust the CI programming. Recently, technology has emerged that permits accurate segmentation of intra-cochlear anatomy and localization of CI electrodes using clinical head CTs in works such as [2,3]. Studies have shown that the use of such image guidance to program the CI leads to a significant improvement in hearing outcomes [4].

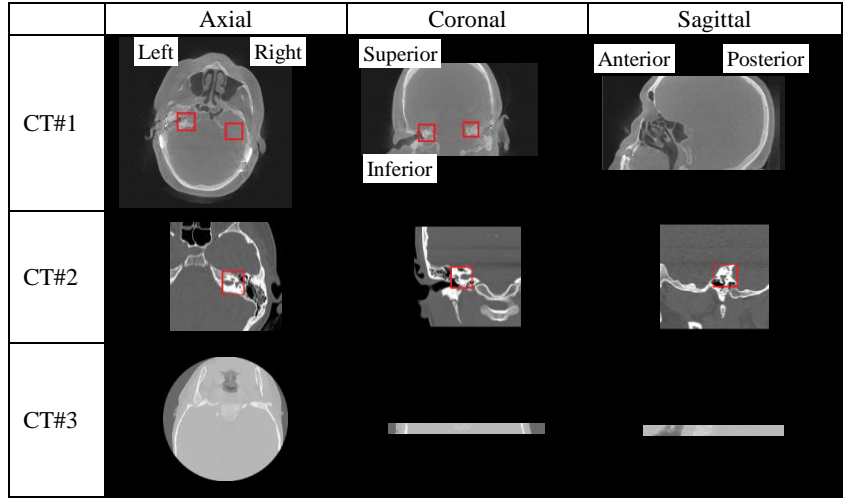

**Fig. 1** Three exemplar CTs from our dataset. The inner ears are shown in red boxes if they are present in the volume and visible in the slice

However, currently the image guidance technology based on the techniques above cannot be fully automated. One hurdle is the heterogeneity of the clinical head CTs. The patients' CT volumes could be obtained from multiple sites. They are acquired using CT scanners of different brands without a standardized imaging protocol and the field of view of the CT volumes varies greatly. Fig. 1 shows three representative examples. Here, CT#1 includes both inner ears. In CT#2, though the right half and a large portion of the left half of the head are present, only the right inner ear is included. In CT#3, only a narrow horizontal portion of the head is imaged and neither inner ear is present.

Because of this heterogeneity, when a new CT volume is received, manual image content documentation, i.e., what ear(s) is/are shown in the volume, and labelling, i.e., where is/are the ear(s) is needed to initialize the subsequent processing steps. Specifically, a technician needs to manually label which inner ear(s) is/are included in the volume and locate it (them). There are several factors that make automating this task challenging: images acquired from different scanners usually have different intensity characteristics and some images have really low quality. Also, the implants could cause serious beam-hardening imaging artifacts. In this paper, we solve this problem using a deep volume-to-volume regression network to directly relate images to probability maps which indicate inner ear positions. We further improve the detection performance by incorporating a false positive suppression strategy at the time of training and applying a post-processing shape-based constraint.

## 2    Methods

### 2.1    Data

|        | Training data |        |       | Test data |        |       |
|--------|---------------|--------|-------|-----------|--------|-------|
|        | w/ CI         | w/o CI | Total | w/ CI     | w/o CI | Total |
| Number | 466           | 2136   |       | 389       | 2097   |       |
|        | Conventional  | Xoran  | 2602  | Conventional | Xoran | 2486 |
| Number | 2022          | 580    |       | 1935      | 551    |       |

**Table 1.** Distributions of our CT data w.r.t. presence of CI and w.r.t. scanner type

The data we use in this study include head CTs from 322 patients. Since acquisitions were obtained both pre-operatively and post-operatively, and multiple reconstructions could be performed for one acquisition, one patient could have several CT volumes. In total, we have 1,593 CT volumes. The CTs are also obtained from both conventional and Xoran xCAT® (denoted as "Xoran") scanners. Xoran scanners are flat-panel, low-dose scanners and images acquired with such scanners typically have a lower quality than those acquired with conventional ones. We label the CT volumes according to the scanner type. The volumes we have also include regions of different sizes and resolutions. The size ranges from 10 mm to 256 mm in the left-right and anterior-posterior directions and from 52 mm to 195 mm in the inferior-superior direction. The resolution varies from 0.14 mm to 2 mm in the left-right and anterior-posterior directions and from 0.14 mm to 5 mm in the inferior-superior direction. We randomly split the data into a training set and a testing set. We make sure one patient's data is not split into different sets. The numbers of CTs in the training set and the testing set are 798 and 795, respectively. For each volume, we visually check the presence of inner ears. For each visible inner ear, a pre-defined landmark point close to the cochlea is manually selected. As we have mentioned, scans could include both inner ears, only one (left/right), or neither. However, the number of image volumes in each of these four categories is not balanced. Indeed, in our current data set about 80% of the volumes include both ears. About 20% include one inner ear. Image volumes that include neither inner ear are rare. To solve this issue, we augment each set by cropping sub volumes from volumes that include both ears to create artificial samples for the other three categories. After cropping, substantial but reasonable deformation, including isotropic scaling, rotation and skewing are applied to increase data variance. All image volumes are resampled to $2.25\times2.25\times2.25$ mm$^3$/voxel and symmetrically cropped or padded to $96\times96\times96$ voxels. In Table 1, we list the number of CT volumes with (w/ CI) and without (w/o CI) implant. We also specify the number of volumes acquired with a conventional scanner and with a Xoran scanner. In the test set, there are 625, 625, 625 and 611 CTs that includes both, left, right and neither ear(s), respectively.

## 2.2 3d U-Net with false positive suppression and a shape constraint

In recent years, methods including [5] and [6] have been proposed to localize organs in CTs using 2D CNNs as a classification tool. These authors use 2D slices in three orthogonal planes extracted from 3-d images volumes to train the CNNs. The location of the organs in the test image is inferred by testing each slice and aggregating them. These models require one forward propagation for each image slice thus requiring hundreds of forward propagations for one volume. Also, the 3-d contextual information is not leveraged. More recently, dense image-to-image or volume-to-volume networks have been favored. In [7], an image-to-image network is proposed to localize landmarks in cardiac and obstetric ultrasound images. The raw 2D image is used as the input and the output is designed to be an action map. In the action map, two vertical lines are drawn which intersect at the position of the landmark, dividing the image into four regions. For a new image, the landmark position can be inferred by aggregating the output action map. This approach only deals with 2D images. In [8], Yang et al. proposed a neural network that can use a whole 3-d CT volume as the input to find a set of vertebra points. However, the unique distribution of many closely-spaced, chain-shaped vertebra points provides far more supervision information in training than what is available in our application.

In this paper, we propose to use a 3d U-Net [9] to map a whole 3-d image volume to two probability maps that have the same dimensionality as the input volumes. As is shown in Fig. 3, the 3d U-Net requires a 3-d volume as input. It consists of multiple convolution-pooling layers which encode the raw input image into low-resolution, highly-abstracted feature maps. Following them are multiple convolution-upsampling layers, which decode the abstracted feature maps into output with the same resolution as the input, in a symmetrical way as the encoding layers. In our first attempt, at the training stage, for each inner ear, we design the probability map as a 3-d Gaussian function centered at the manually labeled landmark position. The standard deviation of the Gaussian is chosen as σ = 3. The maximum is scaled to 1. Any value below 0.05 are set to 0. If the inner ear is not present in the image, all values in the corresponding probability map is set to 0. We treat this volume-to-volume mapping as a regression problem. The weighted mean of voxel-wise squared errors between the output probability maps and the probability maps generated with manually labeled inner ear landmarks is used as the loss function. Larger weights are assigned to voxels with non-zero probabilities. They are sparse but are very important for detection. Specifically, in the output probability map, suppose the number of non-zero entries and zero entries are $N_{none}$ and $N_{zero}$, respectively, the weights associated with none-zero entries and zero entries are $w_{none}$ and $w_{zero}$ defined as follows:

$$\begin{cases} w_{none} = \frac{N_{zero}}{N_{none}+N_{zero}} \\ w_{zero} = \frac{N_{none}}{N_{none}+N_{zero}} \end{cases}. \qquad (1)$$

For each new volume, using the trained network, we generate two probability maps, one for the left ear, the other for the right ear. For each probability map, we find its maximum. If it is larger than $p_{thres}= 0.5$, we predict that the corresponding inner ear is present. Otherwise, we predict that it is absent.

Results we obtained with this approach were not satisfactory because it led to a large number of false positives. We observed that the response map associated with one inner ear could have a very high response at the location of the other ear, possibly due to their similar intensity characteristics. In turn, this led to a substantial number of wrong detections. To solve this problem, we incorporate a false positive suppression strategy during training. Specifically, for the probability map associated with one ear, if the ear on the other side of the head is included in the image, we force the values around this second ear to be negative rather than zero to penalize the detection of the erroneous ear. The negative values that are used are the same Gaussian-distributed values that are used for the correct ear but centered on the incorrect ear and multiplied by minus one. By penalizing the network in such a way, we effectively suppress the number of false positives as will be shown in the results section. We train the neural network using stochastic gradient descent (SGD) with 0.9 momentum and an initial learning rate of 0.0001. The batch size is set to 1. The code is written in Keras [10] on a Nvidia Titan X GPU. The training takes ~2 days. Fig. 3 shows our final network architecture. The response maps of a test image with a left inner ear before and after using false positive suppression are shown in Fig. 4. This figure shows that the false positive caused by the right ear is effectively suppressed.

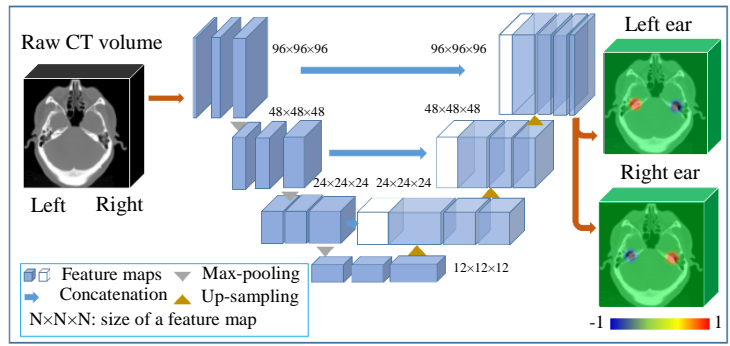

**Fig. 3** Network architecture, input, and output for our approach

Even though the aforementioned method suppresses false positives caused by the contralateral ear, other false positives remain present at other random locations, e.g., the location of the CI transmitters in some post-operative CTs, as shown in Fig. 5. Here, we capture the spatial relationship between inner ear pairs using a low-dimension shape model and use this *a-priori* information to further evaluate the plausibility of the detected inner ear pairs. Specifically, in the training set, we collect the coordinates of the inner ear pairs. For the $i^{th}$ pair, these are denoted as $l_{left}^i = (x_{left}^i, y_{left}^i, z_{left}^i)$ and $l_{right}^i = (x_{right}^i, y_{right}^i, z_{right}^i)$ for the left and right ear, respectively. We subtract from each point the center of the two and stack the coordinate vectors to create a 6-d shape vector $s^i$. The mean shape is computed as

$$\bar{s} = \sum_{i=1}^{N} s^i. \qquad (2)$$

Here, $N$ is the number of inner ear pairs in the training set. The modes of variation of the shapes are computed as the $k$ eigenvectors $\{\vec{u_j}, j = 1,2, ..., k\}$ of the covariance matrix of $\{s^i, i = 1,2, ..., N\}$. The $k$ (in our case, $k=3$ because ears come in pairs) non-zero eigenvalues associated with these are $\{\lambda_j, j = 1,2, ..., k\}$. Suppose the projections of $s^i - \bar{s}$ onto $\{\vec{u_j}, j = 1,2, ..., k\}$ are $\{b_j^i, j = 1,2, ..., k\}$. The Mahalanobis distance between $s^i$ and the mean shape $\bar{s}$ is thus

$$M(s^i, \bar{s}) = \sqrt{\frac{\sum_{j=1}^{k} {b_j^i}^2}{\lambda_j}}. \qquad (3)$$

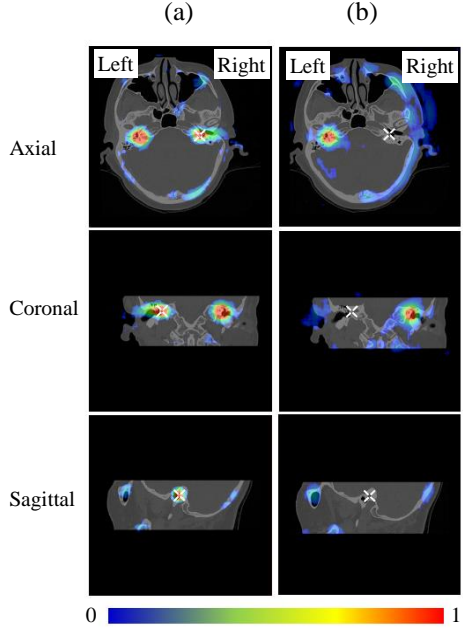
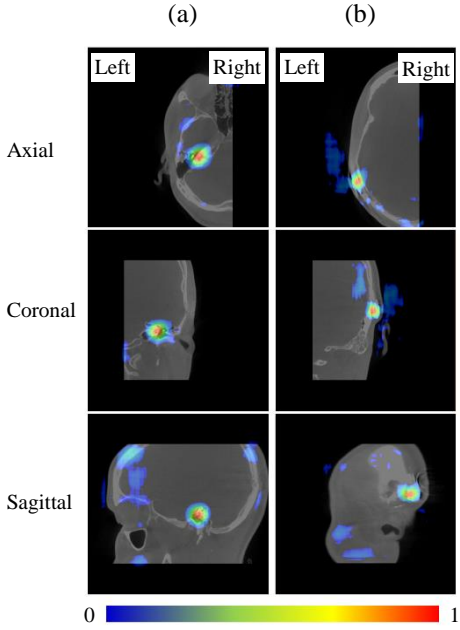

**Fig. 4** The response maps of an input image containing a left ear. Column (a): before applying false positive suppression: the response at the right ear is also very high, (b): after applying false positive suppression, the false positive is eliminated (the location of the right ear is marked).

**Fig. 5** The response maps of an input image that includes the left half of the head. Column (a): the response map associated with the left ear, (b): the response map associated with the right ear. The response at the location of the CI transmitter is so high that it is detected as an ear. It can be eliminated by the shape-based constraint.

It measures how much the spatial distribution of the ear pair deviates from the most 'common' ear pair distribution. We record the maximal Mahalanobis distance of the training shapes as $M_{max}$. For each test volume, when two inner ears are detected from the probability maps, the position vectors of the left and right inner ears, i.e., $l_{left}^{test} = (x_{left}^{test}, y_{left}^{test}, z_{left}^{test})$ and $l_{right}^{test} = (x_{right}^{test}, y_{right}^{test}, z_{right}^{test})$, are stacked and demeaned to create a shape vector $s^{test}$. If $M(s^{test}, \bar{s}) > M_{max}$, we reject the detected inner with the smallest response.

## 3 Results

In Table 3, we show the confusion matrix we have obtained with our testing set. In the upper part of Table 4, we show results with and without the false positive suppression training method and the shape-based constraint. We successfully improve the accuracy by ~7% by using the suppression and the shape-based constraint.

For the test image volumes that are correctly classified, we calculate the localization error, which is shown in the lower part of Table. 4. The localization error is computed as the distance between the manually labeled inner ear position and the automatic localization. For comparison purpose, we use a Random Forest (RF)-based approach [11] developed for head CT landmark localization to find the same landmark in our current dataset and we report the localization accuracy obtained with this method. A number in bold indicates that the localization error generated by the method in this row is lower and significantly different from the other method. The results show that the proposed method produces substantially lower localization error for all image groups. All differences for the five groups of comparisons are statistically different using a paired t-test ($p<0.01$).

| Predict / Truth | B | L | R | N |
|---|---|---|---|---|
| B | 618 | 5 | 1 | 1 |
| L | 0 | 619 | 2 | 4 |
| R | 0 | 4 | 613 | 8 |
| N | 1 | 4 | 5 | 601 |

**Table 3.** Confusion matrix for each category with 'B', 'L', 'R', 'N' indicating volumes including both, left, right and neither ears, respectively.

| | Categorization of error rate | | | | |
|---|---|---|---|---|---|
| | Classified by presence of CI | | Classified by scanner | | Overall |
| | w/ CI | w/o CI | Conventional | Xoran | |
| Before FP reduction | 14.65% | 7.39% | 4.65% | 22.14% | 8.53% |
| After FP reduction | 0.77% | 1.53% | 1.50% | 1.09% | 1.41% |
| Localization error (in mm) | | | | | |
| Proposed | **2.32±2.34** | **2.48±2.35** | **2.41±1.13** | **2.57±4.49** | **2.45±2.35** |
| RF-based | 6.80±18.14 | 5.39±11.80 | 5.01±11.14 | 8.17±19.61 | 5.87±13.57 |

**Table 4.** Categorization of error rate before and after the false positive suppression and shape-based constraint (denoted as FP reduction), and localization error using our proposed method and the baseline RF-based method.

## 4      Conclusions

In this paper, to detect the inner ears in head CTs, we have proposed to use the 3d U-Net to regress the image volume directly to probability maps associated with each inner ear. By incorporating a novel false positive suppression strategy at the time of training and applying a shape-based constraint, we achieve a 98.51% detection accuracy. We achieve a localization error of 2.45mm, which is much better than an RF-based method that has been proposed recently to achieve the same.

**Acknowledgments:**   This work has been supported by NIH grants R01DC014037, R01DC008408, R01DC014462 and Advanced Computing Center for Research and Education (ACCRE) of Vanderbilt University. The content is solely the responsibility of the authors and does not necessarily represent the official views of this institute.